\documentclass[conference, 11pt]{IEEEtran}
\usepackage{amsmath,amssymb,amsfonts}
\usepackage{algorithmic}
\usepackage{graphicx}
\usepackage{textcomp}
\usepackage{xcolor}
\usepackage{graphicx}
\usepackage{listings}
\usepackage[ 
    colorlinks=true,
    pdfborder={0 0 0},
    linkcolor = blue,
    urlcolor  = blue,
    citecolor = blue,
    anchorcolor = blue
]{hyperref}
\usepackage{cleveref}
\mathchardef\mhyphen="2D  

\usepackage[backend=biber, citestyle=authoryear, bibstyle=authoryear]{biblatex}
\usepackage{csquotes}
\bibliography{refs}

\def\BibTeX{{\rm B\kern-.05em{\sc i\kern-.025em b}\kern-.08em
    T\kern-.1667em\lower.7ex\hbox{E}\kern-.125emX}}
\begin{document}

\title{Unveiling Global Discourse Structures: Theoretical Analysis and NLP Applications in Argument Mining}

\author{\IEEEauthorblockN{Christopher van Le}
\IEEEauthorblockA{\textit{\textit{\textit{University of Applied Sciences for Engineering and Economics} } } \\
Berlin, Germany \\
Christopher.Le@Student.HTW-Berlin.de}
}

\maketitle
\thispagestyle{plain}
\pagestyle{plain}
\begin{abstract}
Particularly in the structure of global discourse, coherence plays a pivotal role in human text comprehension and is a hallmark of high-quality text. This is especially true for persuasive texts, where coherent argument structures support claims effectively. This paper discusses and proposes methods for detecting, extracting and representing these global discourse structures in a proccess called Argument(ation) Mining. We begin by defining key terms and processes of discourse structure analysis, then continue to summarize existing research on the matter, and identify shortcomings in current argument component extraction and classification methods. Furthermore, we will outline an architecture for argument mining that focuses on making models more generalisable while overcoming challenges in the current field of research by utilizing novel NLP techniques. This paper reviews current knowledge, summarizes recent works, and outlines our NLP pipeline, aiming to contribute to the theoretical understanding of global discourse structures.
\end{abstract}

\tableofcontents

\section{Introduction}
In persuasive writing, the discourse structures or in other words argumentative structures determine the presence of global coherence. Most studies have focused on what was named logos by Aristotle, i.e., the persuasion by consistency and presentation of evidence according to \textcite{SpeechAndProcessing}. Extracting these features and their components from text is a discipline of computational linguistics. Sentiment analysis, part-of-speech tagging or named-entity recognition solve partial problems of extracting discourse structure as a whole. \textcite{compu-lingu} states, that "the profound analysis of discourse must employ a theory of discourse comprehension and production with which to conduct the analysis". This theory can be can be found in the way arguments are modeled and extracted in argument mining.
\subsection{Argumentative Structures}
\begin{figure}
    \centering
    \includegraphics[width=0.5\textwidth]{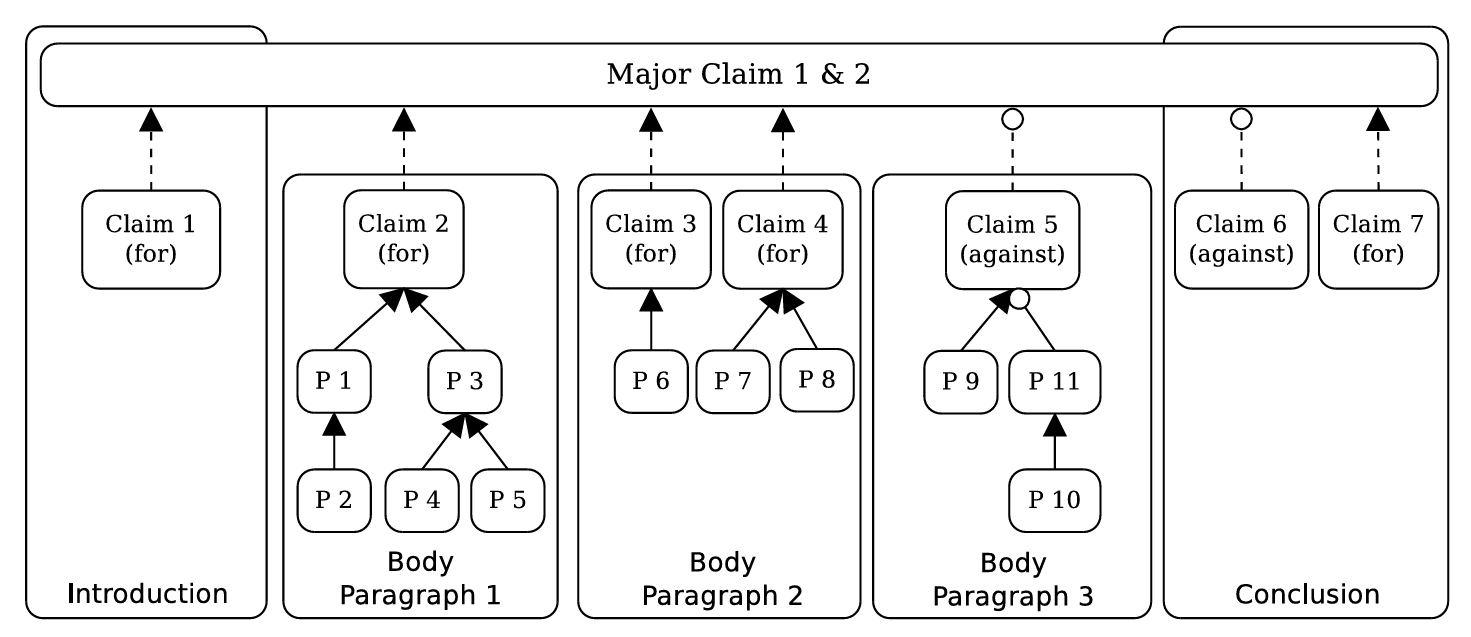}
    \caption{Argument structure from an example essay \parencite{ParsingArgumentationStructures}. An argument modeled in this way can be found in the Appendix A}
    \label{fig:AStructure}
\end{figure}

To represent an argumentative structure, an argument can be broken down into its components and annotated to represent their semantic function. \textcite{stab-gurevych-2014-identifying} and other researchers commonly use a labelling schemes that distinguishe between Claims and Premises. A claim is seen a controversial part of an argument, that needs support by a premise, which provides a reason to believe the claim. Major Claims signify the root of an argument, coherently representing its theme. Other research uses different labels reflecting various levels of logical separation and granularity. Components are usually linked by a classified relation, such as Support or Attack, from the source component to its receiver. Figure \ref{fig:AStructure} depicts an example of an argument structure using the aforementioned annotation scheme. In the diagram, arrow-heads and circles denote support and attack respectivley, while "P" stands for a premise. For this example a tree structure was chosen, but other non-hierarchical representations such as graphs are also common. To formalize the annotation process, annotation frameworks specify the definition of components and relations, and the way in which they are annotated.
\subsection{Argument Mining}
The task of detecting and classifying these features is called Argument(ation) Mining. It is a field of corpus-based discourse analysis and its main objective can be described as predicting "the argument structure from an unstructured text" \parencite{EndToEndAM}. Argument mining is becoming increasingly popular among researchers, partly because of its commercial potential. \textcite{AMSurvey}  argue that opinion mining and sentiment analysis, two NLP tasks, have been successful in fields such as marketing, public relations, and financial market prediction. These technologies have contributed to a market worth around \$10 billion. However, they point out that these methods do not take into account the argumentative structure of texts and only reveal expressed opinions, not the underlying reasons for them. Argument mining has an advantage over established techniques by bridging the gap between identifying opinions and extracting reasoning. For instance, companies that work with textual customer data, such as feedback on products and services, might want to comprehend the reasoning behind an opinion.

In order to extract an argument structure from text \textcite{ParsingArgumentationStructures} propose three sequential tasks:
\begin{enumerate}
  \item Span identification
  \item Component classification
  \item Relation classification
\end{enumerate}
\begin{figure}
    \centering
    \includegraphics[width=0.5\textwidth]{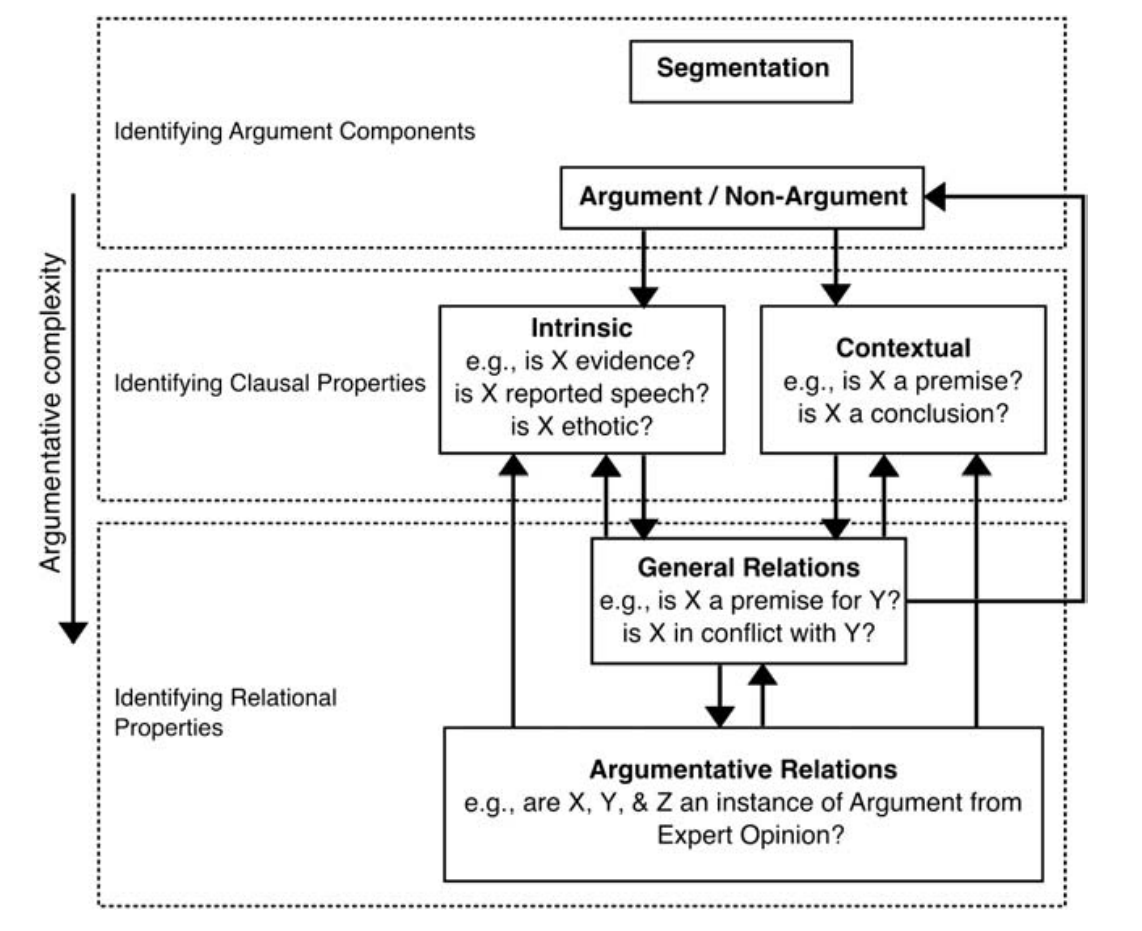}
    \caption{Visualization of complexity levels of tasks in argument mining techniques \parencite{AMSurvey}.}
    \label{fig:AMComplexity}
\end{figure}
This description of a workflow does not effectively convey the difficulty involved. In Figure \ref{fig:AMComplexity} an increasing complexity of tasks can be observed. It is positively correlated with the order of the steps. The identification of argumentative components is the first step in argument mining and is considered least complex. The last step, the identification of relational properties, emerges as the most complex. This cascade of complexity amplifies the challenge of creating a high-performing argument mining pipeline. The absolute performance of each step depends on that of a previous step. For example, the outcome of a relational classification is highly dependent on the correct span or type identification of a component. If a range is incorrectly identified, subsequent classifications based on that range will be incorrect. A way to avoid these cascading errors is described by \textcite{TowardsRelationBasedAM}: instead of using the aformentioned sequence of tasks, they propose a relation-based approach, arguing for classifying relations first, and then considering propositions to be argumentative if they have a relation connecting them.

\begin{figure}
    \centering
    \includegraphics[width=0.45\textwidth]{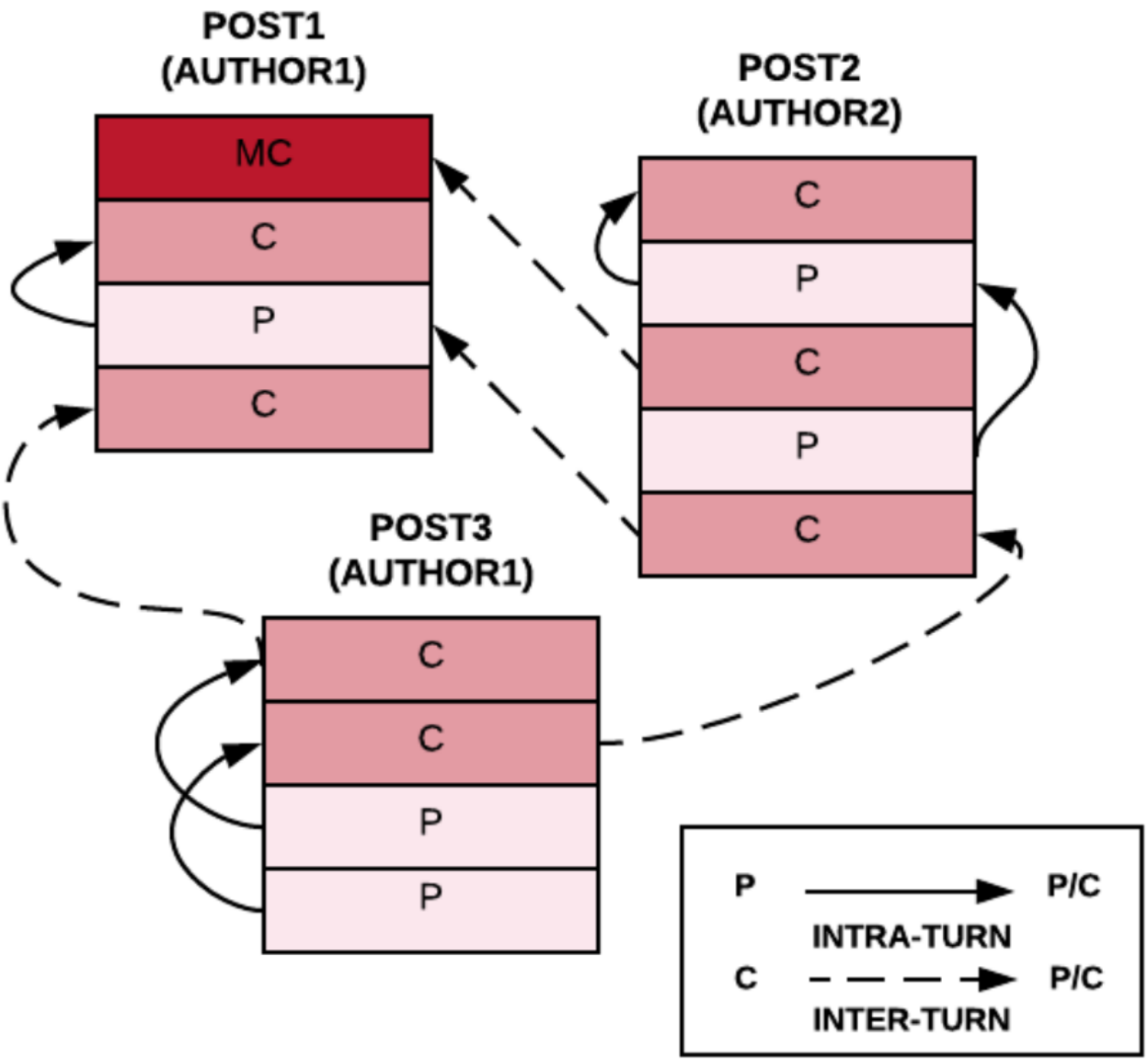}
    \caption{Multilogue argument structure \parencite{Ampersand}.}
    \label{fig:Ampersand}
\end{figure}

While the term argument mining can be used to characterise individual sub-tasks, this paper will refer to it as the proccess of identifying components, classifying them and classifying their relations. In order to facilitate all of these steps, advanced natural language processing techniques are used to create sophisticated argument mining models. This usually involves training a machine learning model with annotated corpora as training data. Typically for these models the amount of training data has to be high for a well-performing model. \textcite{EndToEndAM} highlight the absence of standardised annotation frameworks and the resulting lack of uniform training data. Additionally, they mention the challenge posed by varying argumentative styles in certain domains. Even when limiting the corpus to domain-specific language, the results remain unsatisfactory, particularly in relation classification \parencite{TransformerHealthcareAM}. In a different approach by \textcite{EndToEndAM}, the researchers propose a technique called 'Multi-Task Argument Mining' to address the problem of non-standardized annotation frameworks and the lack of corpora suitable for training data. They achieve a more flexible model by creating an architecture, that allows to "handle different concepts of spans, types of component labels, and graph structures". The majority of argument mining research has focused on monologues. Exceptions to this rule include works that model multilogue, such as \textcite{Ampersand}, in which user discourse on internet forums is mined. This paper proposes a model which considers that argumentation is not only taking place on the micro-level, but also on the macro-level, refering to argumentation as as process. Figure \ref{fig:Ampersand} shows the inter-turn relationships that connect the argumentative units with each other. The diagram illustrates how the arguments of two entities interact on a component level.

So far there has not been presented a general purpose argument mining model and it is unclear if the task can be generalized to such extent taking into account the variability of argumentative styles. \textcite{AMSurvey} conclude that argument mining is a challenging task and that it is important to note that the information required to comprehend an argument is not limited to what is explicitly stated.

\section{Concepts}
\subsection{Corpora}
The creation of corpora, a fundamental resource in the field of argument mining, is a complex and challenging process. This is particularly noticeable when examining the limited number of corpora, that annotate argumentative structure. \textcite{ParsingArgumentationStructures} state, that existing corpora lack important qualities such as including non-argumentative text, annotating claims and premises, and indicating the reliability of annotations. The primary reason for this limitation is the requirement for manual text annotation. In natural language, categories such as claims, premises, and their interrelationships are not precisely defined and are susceptible to the annotator's subjective judgement. In order for a corpus to be considered of high quality, a well-documented annotation process, the involvement of multiple annotators and a disclosed Inter-Annotator Agreement (\textbf{IAA}) are essential. The IAA is a statistical measure that describes the degree of consistency among multiple annotators in making the same annotation decision for a given label category or class. These neccessary measures are resource-intensive to provide and make the creation of corpora a significant undertaking.

 An indication of this can be observed in a survey by \textcite{AMSurvey}, investigating most of the significant argument mining corpora. It shows that 7 out of 16 have either a single annotator or do not report their IAA. The remaining studies typically use either Cohen's kappa or Krippendorff's alpha to measure overall agreement. Additionally, some studies differentiate between agreement on component types and relations, using these statistical methods separately for each. This allows for a thorough understanding of their annotation results. Missing agreement on a standardized and ubiquitous guideline for annotation, the field of research struggles to maintain consistency and comparability across different corpora. This hinders the progress and reliability of linguistic and computational studies and narrows down the scope and generalizability of the resulting work. To address these issues, projects such as the Argument Interchange Format Database \parencite{AIFdb} are working towards standardization and exchange of annotated arguments using the Argument Interchange Format. This involves providing a platform for storing and accessing argument data in a uniform way, helping to unify corpus building efforts.

\begin{figure}
    \centering
    \includegraphics[width=0.5\textwidth]{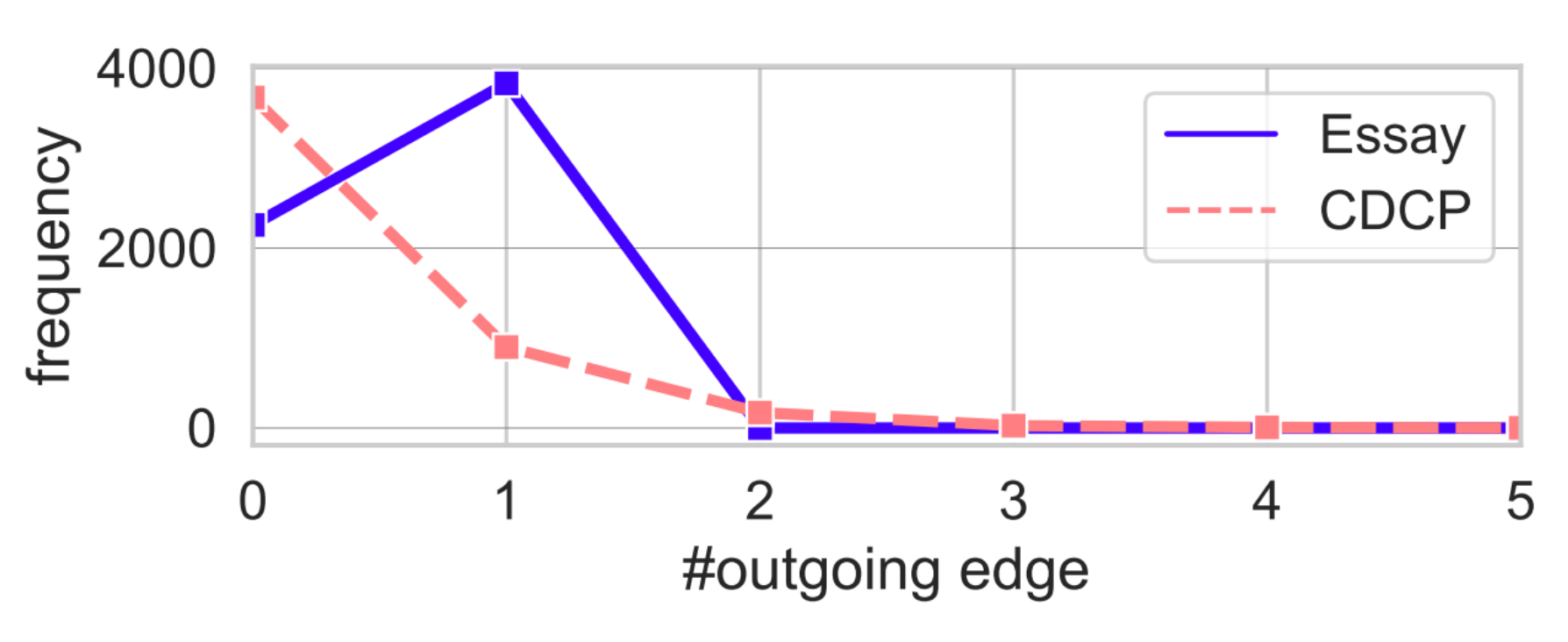}
    \caption{Distribution of outgoing edges from nodes in CDCP and Essay \parencite{TowardsNonTree}}
    \label{fig:OutgoingEdges}
\end{figure}

The linguistic characteristics of corpora present an additional challenge to their use. Figure \ref{fig:OutgoingEdges} illustrates the differences in features between specific types of text. \citeauthor{CDCP}'s (\citeyear{CDCP}) Consumer Debt Collection Practices (\textbf{CDCP}) corpus contains user-generated comments on lawmaking regarding consumer debt collection practices by the Consumer Financial Protection Bureau, annotated as graphs. The Essay corpus by \textcite{ParsingArgumentationStructures} consists of essays annotated as trees. A comparison was made between the distribution of outgoing relations from tree-components in Essay and the edges of nodes from CDCP, revealing a significant discrepancy. This suggests different styles of argumentation consist of different argumentative structures. In this case CDCP contains more isolated propositions than Essay, which in turn has more singular linkage of arguments. These inherent differences make generalisation more challenging.

\subsection{Automatic Argument Structure Extraction}
The entities that comprise argrument structure in text can be formalized like this in reference to \textcite{EndToEndAM}: spans \((S)\), components \((C)\), relations \((R)\), with \(\langle s,e\rangle \in (S)\) denoting span start \(s\) and span end  \(e\). \(\langle s,e, c\rangle \in (C)\) adding the component type  \(c\) to \((S)\). \( \langle s_{src},e_{src},s_{tgt},e_{tgt}, r\rangle \in (R)\) with \(s_{src}\) and  \(s_{tgt}\) representing the source- and targets-side span, while \(r\) defining the type of relation.

\begin{figure}
    \centering
    \includegraphics[width=0.5\textwidth]{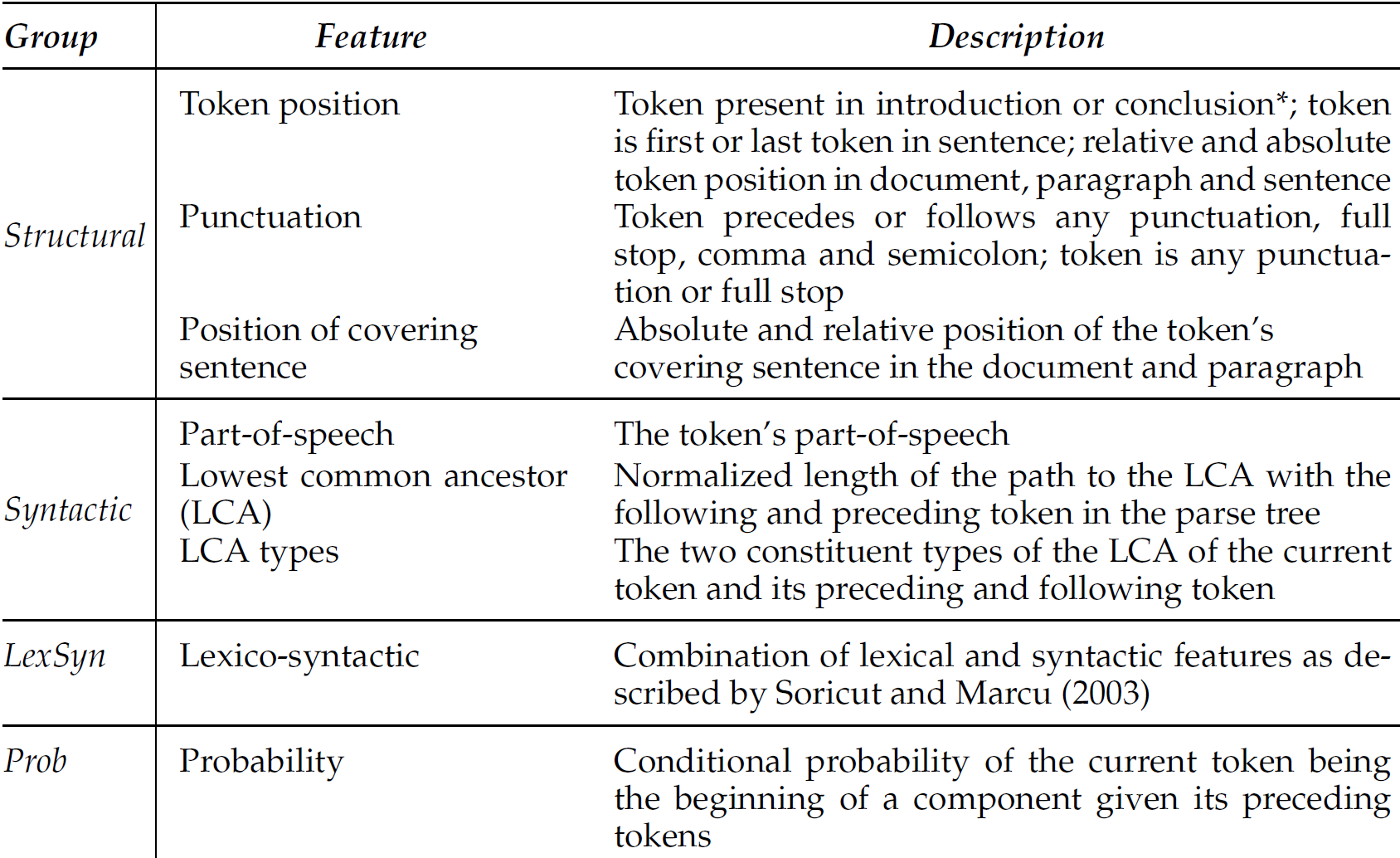}
    \caption{A subset of features used by \textcite{ParsingArgumentationStructures} for component classification }
    \label{fig:ClassFeatureSubset}
\end{figure}

Assuming a linear task sequence a typical systematic process could begin by identifying \((S)\). To do so, the workflow would start with tokenization, the step in which text is broken down into individual words or tokens. This is foundational to any text analysis task as it transforms unstructured data into a structured format. Next, the token is encoded, typically using embeddings, which are mathematical representations of words in a high-dimensional space, that can embed lexical, structural, contextual and syntactic features. Transformer embeddings have been found to be effective in doing so \parencite{TransformerHealthcareAM}. Before embeddings could capture a comprehensive meaning representation of text, a wide range of characteristics would be used as individual feature inputs to models, instead of being integrated into the embedding. \textcite{ParsingArgumentationStructures} tried to contextualise the input to their model by extracting features from each token and their surroundings. The aforementioned features were employed as inputs for a Conditional Random Field (CRF) machine learning model. This model was designed to take advantage of its ability to consider "neighbouring" samples during component identification, effectively accounting for local discourse structures. An example of features used in this work can be found in Figure \ref{fig:ClassFeatureSubset}.
This step allows the model to understand the semantic meaning of words and their relationships with each other. Following this, the workflow involves identification of \((S)\) and classification of \((S) \rightarrow (C)\). This step is responsible for identifying the argumentative components in the text, such as claims and premises, and classifying them into their respective categories. The fourth step, link detection, establishes the relationships between these argumentative components, determining which claims are connected to which premises: \( \langle s_{src},e_{src},s_{tgt},e_{tgt}\rangle \). Finally, the relation label classification assigns labels  (\(r\)) to these relationships, providing further nuance to how the identified components interact. This could include labels such as 'supports', 'refutes' or 'undermines'.  A function that feeds a classifier with two components that have been identified as related might look like this:
\[P(r_{i\rightarrow j}) = f_{CLASS}(C^{(tgt)}_j, C^{(src)}_i)\]
where \(P(r_{i\rightarrow j})\) is the probability distribution of the relation-label classes. 

\section{Related Work}
In recent years, argument mining has become a productive field of research. Although limited to a small number of corpora, various approaches have been developed for the automatic extraction of argument structures. This section will explore the methods employed to extract arguments from text and evaluate their advantages and disadvantages. The selected works were chosen based on their relevance, the reliability of their findings, and the range of technologies utilised.
\subsection{ Integer Linear Programming (\textbf{ILP}) Joint Model } \label{sec:ILP}
\begin{figure}
    \centering
    \includegraphics[width=0.5\textwidth]{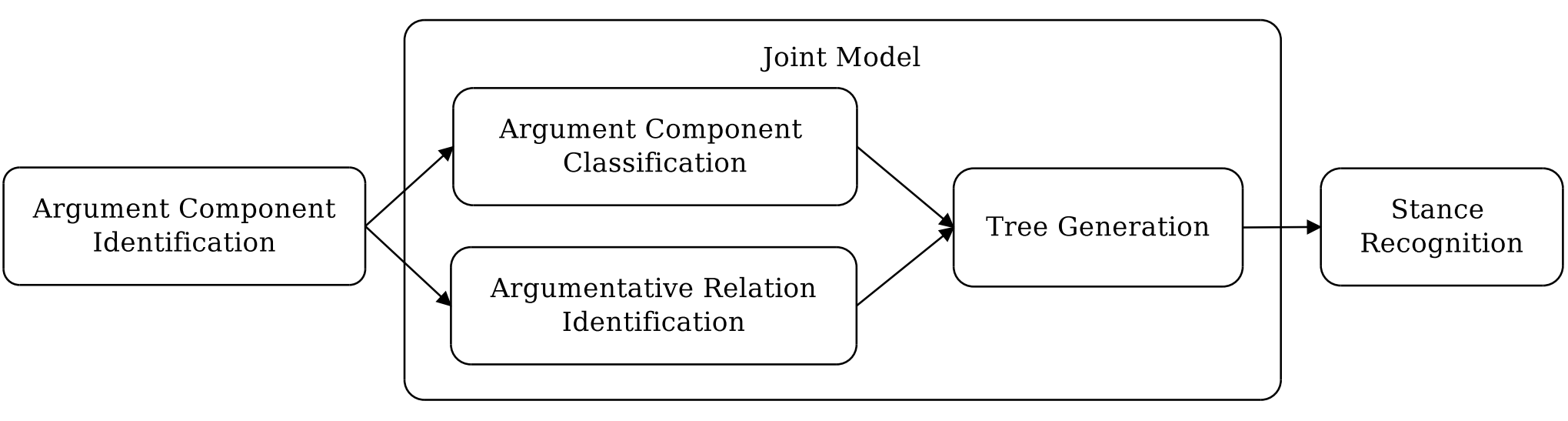}
    \caption{Task architecture of an argumentation structure parser \parencite{ParsingArgumentationStructures}}
    \label{fig:IlpJointModel}
\end{figure}

\textcite{ParsingArgumentationStructures} introduce a corpus of persuasive essays annotated with argumentation structure in their paper. For the corpus they provide numbers on the Inter-Annotator Agreement (\textbf{IAA}), reporting an overall agreement of \(\alpha_U = 0.767\). Taking into account the IAA's of their previous work \parencite{stab-gurevych-2014-identifying}, they tentatively conclude, "that overall human annotators agree on the argument components in persuasive essays" \parencite{ParsingArgumentationStructures}. Additionally, they present a parser for argumentation structure, which uses the corpus for training and validating a machine learning model.  For their parser they propose an architecture shown in Figure \ref{fig:IlpJointModel}. They use a directed sequential approach with a joint model to classify component types and identify their relations. The authors argue that their model is more effective than identifying argumentative relations and components individually using the base classifiers, as their model can more reliably link premises. They have also defined a ruleset to converge the results into a tree structure. In the final step, they classify the argumentative relations. This process considers pairs of components that are linked in the tree structure for classification.

To fit and evaluate their model they opted for a 5-fold cross-validation on their own Essay corpus consisting of 6089 argument components. As their main performance metric they use the macro F1 score. It is a popular metric for evaluating binary classification models and represents the harmonic mean of precision and recall over all labels/classes. They report achieving significant improvements over their heuristic baseline models. These are the resulting macro F1 scores for the individual classification sub-tasks with their ILP joint model: 
\begin{itemize}
    \item Components: 0.826
    \item Relations: 0.751
    \item Stance Recognition: 0.680 
\end{itemize}
For the component identification model they use a CRF model. The resulting component spans are then fed into their ILP model, which combines the use of two Support Vector Machines in conjunction, to recognize the argumentation structure. According to one of their earlier works, the reason for this is that argumentative types and relations share information \parencite{stab-gurevych-2014-identifying}. For instance, the probability of an outgoing relationship is lower for a component classified as a claim than for a premise. That is why an independent approach to those problems holds less potential for good results. In their ILP model the authors present a method for determining the weights of argumentative relations in an adjacency matrix. Claim scores are calculated based on predicted relations, with higher weights assigned to relations pointing to claims and weights set to 0 for relations pointing to premises. This method considers predicted relations, claim scores, and component types to accurately predict argumentative relations.

\subsection{Transformer-Based Argument Mining } \label{sec:Transformer}
\begin{figure*}
    \centering
    \includegraphics[width=0.8\textwidth]{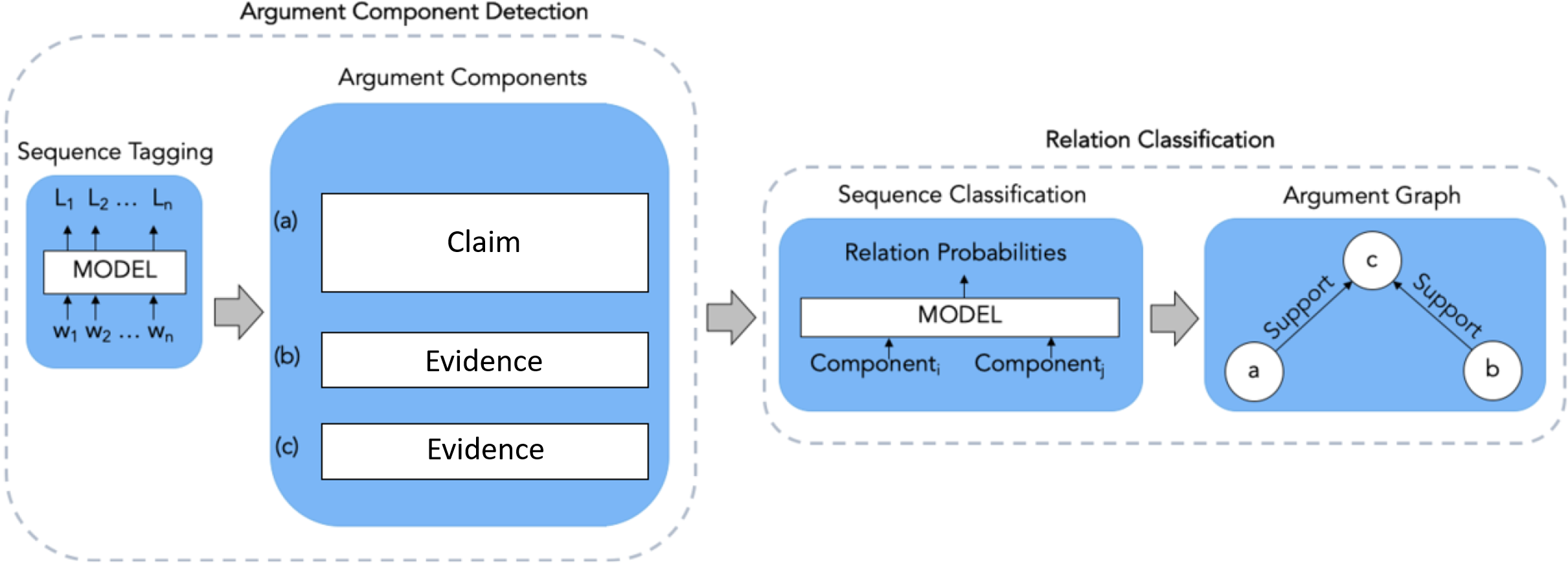}
    \caption{Pipeline of clinical trial argument mining \parencite{TransformerHealthcareAM}. }
    \label{fig:TransformerPipe}
\end{figure*}
In their paper \textcite{TransformerHealthcareAM} focus on mining argumentation in abstracts of Randomized Control Trials (\textbf{AbstRCT}). They extended a corpus from the MEDLINE database, totaling a number of 4198 annotated argument components and 2601 relations. For measuring the IAA, a sample of 30 abstracts from the corpus was drawn. According to the authors, Fleiss' kappa was 0.72 for component annotation and 0.62 for relation annotation, demonstrating moderate to substantial agreement between annotators. They presented an argument mining pipeline that uses neural networks in combination with transformer-embeddings. The neural architectures used in this study include Long Short-Term Memory (\textbf{LSTM}) networks, Gated Recurrent Unit (\textbf{GRU}) networks, and CRFs with various types of embeddings. These range from static, context-insensitive embeddings such as \textbf{GloVe} \parencite{glove}, \textbf{fastText} \parencite{bojanowski2016enriching}, and \textbf{BPEmb} \parencite{bpemb} to dynamic embeddings like \textbf{ELMo} \parencite{ELMo}, \textbf{FlairPM} \parencite{flair}, and \textbf{BERT} \parencite{bert}. The main difference between these two categories is that dynamic embeddings contextualize not only a single word but the entire input sequence. It is important to note that the researchers also experimented with more fine-tuned versions of BERT: \textbf{BioBERT} \parencite{biobert} and \textbf{SciBERT} \parencite{scibert} are transformers specifically trained on biomedical texts and scientific papers respectively. Figure \ref{fig:TransformerPipe} illustrates their architecture, which combines the detection and classification of components into a single task, called multi-class sequence tagging. Tokens \textbf{w} are tagged with a modified Beggining-Inside-Outside-tagging (\textbf{BIO-tagging}) scheme using five labels, B-Claim, I-Claim, B-Evidence, I-Evidence and Outside. Adjacent tagged tokens \textbf{L} can be directly converted to their corresponding component, shown as \textbf{a}, \textbf{b} and \textbf{c}. E.g. this sequence of tokens \(B \mhyphen Claim \rightarrow I \mhyphen Claim \rightarrow I \mhyphen Claim \rightarrow Outside\) would represent a claim. This departs from the conventional sequential pipeline and suggests a more end-to-end oriented approach by combining multiple tasks in a single model. The relationships between the extracted components from the tagged tokens, are then classified by a second model. After that a graph is constructed, reflecting the argument structure of the text.

The results of the sequence tagging task show a clear performance advantage of the dynamic embeddings. The combination of fine-tuned BioBERT embeddings with GRU networks and CRFs resulted in an overall macro F1 score of 0.91. The model's performance was then tested for relation classification, where the use of SciBERT embeddings achieved a macro F1 score of 0.69.

\subsection{Multi-Task Argument Mining (\textbf{MT-AM})}
\begin{figure*}
    \centering
    \includegraphics[width=0.8\textwidth]{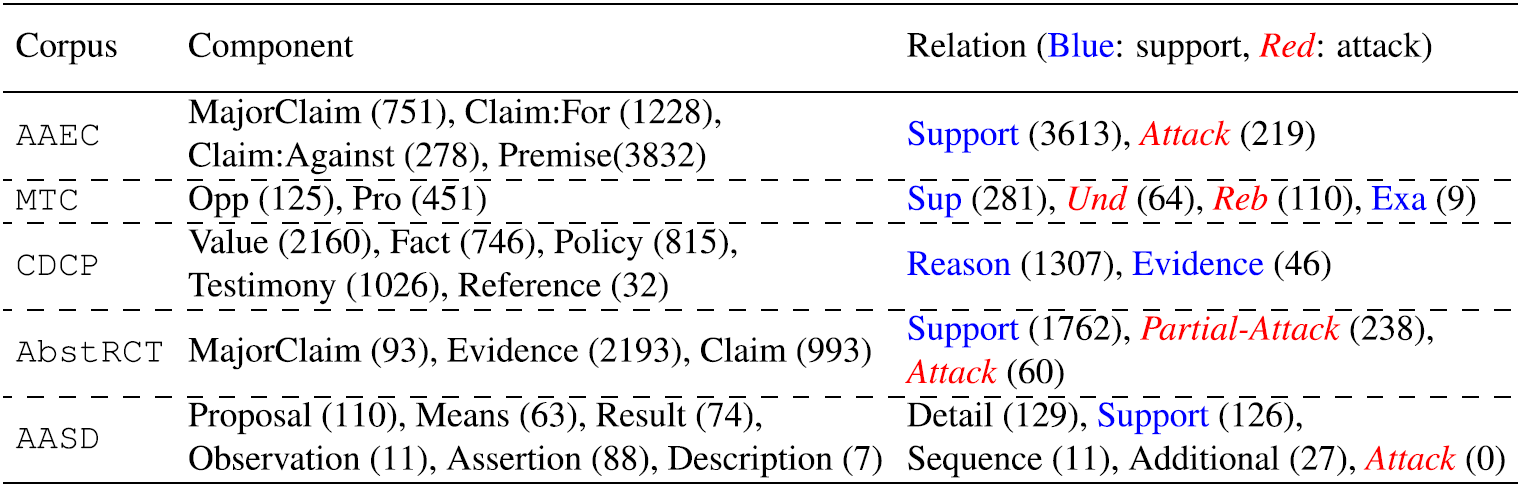}
    \caption{A table of the various corpora used by \textcite{EndToEndAM}}
    \label{fig:E2ECorpora}
\end{figure*}

In their work, \textcite{EndToEndAM} try to address the issue of shortage of training data by proposing a method called Multi-Task Argument Mining (MT-AM). The approach utilises various corpora with different annotation schemes, structures (graph/tree-based) and domains to enhance overall argument mining performance. By leveraging similar argumentation patterns across multiple corpora, they hope, that their model's overall performance can be improved. An overview of the used corpora can be seen in Figure \ref{fig:E2ECorpora}, where AAEC refers to the Essay corpus from \textcite{ParsingArgumentationStructures} in section \ref{sec:ILP} and AbstRCT is the non-extended version of the corpus used by \textcite{TransformerHealthcareAM} in section \ref{sec:Transformer}. Their thorough analysis of the corpora revealed similar characteristics of the annotations and argumentative structure, indicating their potential usefulness in MT-AM. This for example includes the almost universal distinction of relations in support and attack, that can also be seen marked in blue and red in Figure \ref{fig:E2ECorpora}.

The two-staged MT-AM model incorporates as first stage a pre-training phase using multiple auxiliary corpora. This approach enhances the model's ability to recognise transferrable linguistic structures among corpora, aiding its generalisation across tasks. The model's significant feature is its adoption of end-to-end learning. Unlike transition-based parsing methods, it does not require the mapping of component and relation labels on the bases of semantic commonality. Instead, it is trained to recognize these labels directly from the data. During the fine-tuning phase, the model is trained again using the defined target corpus and different hyperparameters, such as a higher loss weight during learning. This is done to prevent performance degradation by supressing distant information from the auxiliary corpora. The model is then ready to be applied to text, preferably with an argument structure similar to that of the target corpus.

The pipeline parsing routine consists of the following steps: The Longformer \parencite{beltagy2020longformer}, a transformer model, is used to generate sequence embeddings. It is specifically designed to handle long text sequences, making it ideal for argument mining tasks, that require understanding of extensive context. After generating the sequence embedding, the next step is to identify and classify spans using a Multilayer Perceptron (\textbf{MLP}) neural network \parencite{nn_foundation}. After identifying and classifying the spans, the sequence embedding undergoes an average pooling operation to transform it into a fixed-size embedding. This enables subsequent layers of the model to process the embedding, regardless of the original sequence length. Finally, a biaffine classifier is used to detect links and classify labels. The biaffine classifier is a type of classifier that models complex relationships between different parts of the data, making it particularly effective for tasks, that are dependent on each other like link detection and label classification. 

As a baseline for evaluation they use, their Single Task (\textbf{ST}) model, which is only trained on the target corpus. Generally the MT-AM pre-trained on all auxiliary corpora and fine-tuned on the target corpus outperformed the ST models. In a comparison with other models, including the ILP Joint model from \textcite{ParsingArgumentationStructures} and a transformer-based approach similar to \textcite{TransformerHealthcareAM}, their ST model performed on par with the competition. It managed to outperform the ILP Joint model on the Essay corpus with a macro F1 score of 0.868 at component classification \parencite{mtc} 

\subsection{Model discussion}
\citeauthor{ParsingArgumentationStructures}'s (\citeyear{ParsingArgumentationStructures}) joint model approach already hints to a development, that can be seen in the following works discussed in this section. They shift away from designated sub-tasks, to a model integrated approach. Even though they succeed in creating a well-performing parser for their Essay corpus, their feature-based approach has shown its limits. Later studies improve on capturing context more intuitively, with for example transformer embeddings, instead of single structural, lexical or syntactical features.  Another notable feature of the presented work is the end-to-end learning of \textcite{EndToEndAM}. They have been able to demonstrate the effectiveness of multitask argument mining integrating various corpora into one model. This highlights two characteristics in the field of argument mining: argumentative structures in different domains contain overlapping information, that can be exploited; models are likely to perform well, when they are specifically trained on their tasks target domain. This claim is also supported by the results of \textcite{TransformerHealthcareAM}. By using a domain-specific transformer, embeddings are created that better represent the context of the domain. This allows machine learning models working with these embeddings to learn argumentative patterns in the language more effectively.
\section{Theoretical Architecture}
\subsection{Requirements}
The aim of this chapter is to present a theoretical pipeline for argument mining that can parse monological persuasive text, regardless of the domain, into its argumentative structure. This involves extracting components and relations and representing them in an indiscriminate annotation scheme. With a reasonably well-annotated and extensive corpus, the model should be able to learn its inherent argumentative structure and perform well in extracting it. To be considered well-performing, the model must accurately distinguish between spans of non-argumentative text and argumentative text, correctly identify and classify argument components and their relations, and represent the argument as a graph-based structure as the return value of its prediction function.
\subsection{Architecture}
To achieve a degree of domain independence, MT-AM uses multiple corpora in its training. Although argument mining corpora are scarce, corpora used for its sub-tasks, such as relation classification, are more readily available. A model that is able to take advantage of an multi-facetted learning approach could be a text-to-text (\textbf{T2T}) transformer. These models have been proven to handle a wide variety of NLP problems, while providing good performance and easy model management. The T5X-model from \textcite{text2text}, offers pre-trained models and built-in functions for fine-tuning. T2T transformers that heavily rely on embeddings. A domain-specific pre-trained transformer, such as SciBERT, may provide benefits in representing structures from scientific text more accurately than a general-purpose transformer. As embeddings serve as the internal intermediate language in T2T transformers, having a precise mathematical representation of the subject matter could be highly advantageous.

In conclusion, the proposed architecture would feature a T2T transformer model using domain-specific embeddings and be trained on argument mining tasks as well as argument mining related subtasks. A deployed application of this pipeline would include a pre-classification of the problem to determine which embedding space (or transformer) to use, depending on the task domain. This could be achieved by calculating the overlap of the input text's embedding with the domain-specific embedding spaces. Since T2T transformers inputs and outputs are text, an interaction with the model, depending on how it was trained, could look like this (used example argument can be found in the appendix):

\textbf{Input}: 
\begin{lstlisting}
    argument mine(ANN-Format): 
    "We should attach more importance..."
\end{lstlisting}

\textbf{Output}: 
\begin{lstlisting}
    T1	MajorClaim 1 512:
    "We should attach more importance..."
    
    T2	Claim 591 714:
    "through cooperation, children can..."
    
    R1	supports Arg1:T2 Arg2:T1
\end{lstlisting}

The T5-model, the predecessor of the T5X, already has been tested in the field of argument mining by \textcite{t2t-AM}. They showcase a significant improvement over many other models, including the ILP Joint model from \textcite{ParsingArgumentationStructures} and the ST model from \textcite{EndToEndAM}. Although the T5-model allows training from scratch, pre-training it on a domain-specific database like SciBERT is a resource-intensive task. To create a nearly general-purpose argument mining model, hundreds of pre-trained models would be required. For reference, the number of parameters used by \textcite{t2t-AM} in their best performing T5-XXL model was 11 billion. By comparison, the ST model had only 149 million. In their paper, they conclude that the limitations of their model were imposed by the memory of their GPU. It remains to be proven whether a T2T transformer model can solve argument mining problems efficiently and effectively.

\section{Conclusion}
In conclusion, we have reviewed the methods for detecting, extracting and representing global discourse structures in a process known as argumentation mining. We have defined key terms and processes of discourse structure analysis, summarised existing research, and identified trends in current methods for extracting and classifying argument components. Our work contributes to the theoretical understanding of extracting global discourse structures by highlighting key characteristics of recent developments in argument mining. We have also shed light on the challenges and complexities involved in argument mining, such as the the sequential nature of the task and the variability of argumentative styles across domains. We have discussed the potential of Multi-Task Argumentation Mining as a technique to address the problem of non-standardised annotation frameworks and the lack of corpora suitable for training data. In our research, we have outlined an NLP architecture for argumentation mining to facilitate the detection and representation of global discourse structures with novel technologies. While we have made significant progress in understanding and representing global discourse structures, it is important to acknowledge that the task of argument mining remains challenging. The information required to understand an argument is not limited to what is explicitly stated and the variability of argument styles across domains poses a significant challenge. Corpora are still scarcely available and universal agreement over annotation frameworks has yet to be achieved. However, based on current developments in the advancement of NLP applications and techniques, especially transformer models, we remain optimistic about the potential of argument mining. 
\printbibliography

\onecolumn
\appendix
\setcounter{secnumdepth}{0}
\textbf{Appendix A: Annotated Argument from \textcite{ParsingArgumentationStructures}}
\begin{center}
\begin{minipage}[c]{\textwidth}

\begin{lstlisting}
T1	MajorClaim 503 575	we should attach more importance to cooperation during primary education
T2	MajorClaim 2154 2231	a more cooperative attitudes towards life is more profitable in one's success
T3	Claim 591 714	through cooperation, children can learn about interpersonal skills which are significant in the future life of all students
A1	Stance T3 For
T4	Premise 716 851	What we acquired from team work is not only how to achieve the same goal with others but more importantly, how to get along with others
T5	Premise 853 1086	During the process of cooperation, children can learn about how to listen to opinions of others, how to communicate with others, how to think comprehensively, and even how to compromise with other team members when conflicts occurred
T6	Premise 1088 1191	All of these skills help them to get on well with other people and will benefit them for the whole life
R1	supports Arg1:T4 Arg2:T3	
R2	supports Arg1:T5 Arg2:T3	
R3	supports Arg1:T6 Arg2:T3	
T7	Claim 1332 1376	competition makes the society more effective
A2	Stance T7 Against
T8	Premise 1212 1301	the significance of competition is that how to become more excellence to gain the victory
T9	Premise 1387 1492	when we consider about the question that how to win the game, we always find that we need the cooperation
T10	Premise 1549 1846	Take Olympic games which is a form of competition for instance, it is hard to imagine how an athlete could win the game without the training of his or her coach, and the help of other professional staffs such as the people who take care of his diet, and those who are in charge of the medical care
T11	Claim 1927 1992	without the cooperation, there would be no victory of competition
A3	Stance T11 For
R4	supports Arg1:T10 Arg2:T11	
R5	supports Arg1:T9 Arg2:T11	
R6	supports Arg1:T8 Arg2:T7	
\end{lstlisting}
\end{minipage}

\end{center}
\end{document}